\documentclass{article} 
\usepackage{iclr2016_conference,times}
\usepackage{url}

\usepackage{natbib}
\usepackage{graphicx}
\usepackage{amsmath}
\graphicspath{{figures/}}
\def\ie{{\it i.e.}}
\def\btheta{{\mathbf{\theta}}}
\def\bX{{\mathcal{X}}}
\def\by{{\mathbf{y}}}
\def\bone{{\mathbf{1}}}

\usepackage{multirow}
\usepackage{ctable}
\usepackage{psfrag}
\newcommand{\norm}[1]{\left\lVert#1\right\rVert}

\title{Manifold Regularized Deep Neural Networks using Adversarial Examples}

\author{Taehoon Lee, Minsuk Choi, and Sungroh Yoon\\
Department of Electrical and Computer Engineering\\
Seoul National University\\
Seoul 151-744, Korea\\
\texttt{\{taehoonlee,minsuk0523,sryoon\}@snu.ac.kr}
}

\begin{document}

\maketitle

\begin{abstract}
Learning meaningful representations using deep neural networks involves designing efficient training schemes and well-structured networks. Currently, the method of stochastic gradient descent that has a momentum with dropout is one of the most popular training protocols. Based on that, more advanced methods (\ie, Maxout and Batch Normalization) have been proposed in recent years, but most still suffer from performance degradation caused by small perturbations, also known as \textit{adversarial examples}. To address this issue, we propose \textit{manifold regularized networks} (MRnet) that utilize a novel training objective function that minimizes the difference between multi-layer embedding results of samples and those adversarial. Our experimental results demonstrated that MRnet is more resilient to adversarial examples and helps us to generalize representations on manifolds. Furthermore, combining MRnet and dropout allowed us to achieve competitive classification performances for three well-known benchmarks: MNIST, CIFAR-10, and SVHN.
\end{abstract}

\section{Introduction}

Deep neural networks have been used successfully to learn meaningful representations on a variety of tasks~\citep{Hinton06,Krizhevsky12}. By adding more hidden units or layers, we can learn more complicated relationships between sensory data and desired outputs. However, increasing the model complexity incurs an enormous solution space. Thus, regularization techniques have been normally combined with deep neural networks to obtain acceptable solutions.

\begin{figure}
\centering
\includegraphics[width=.94\linewidth]{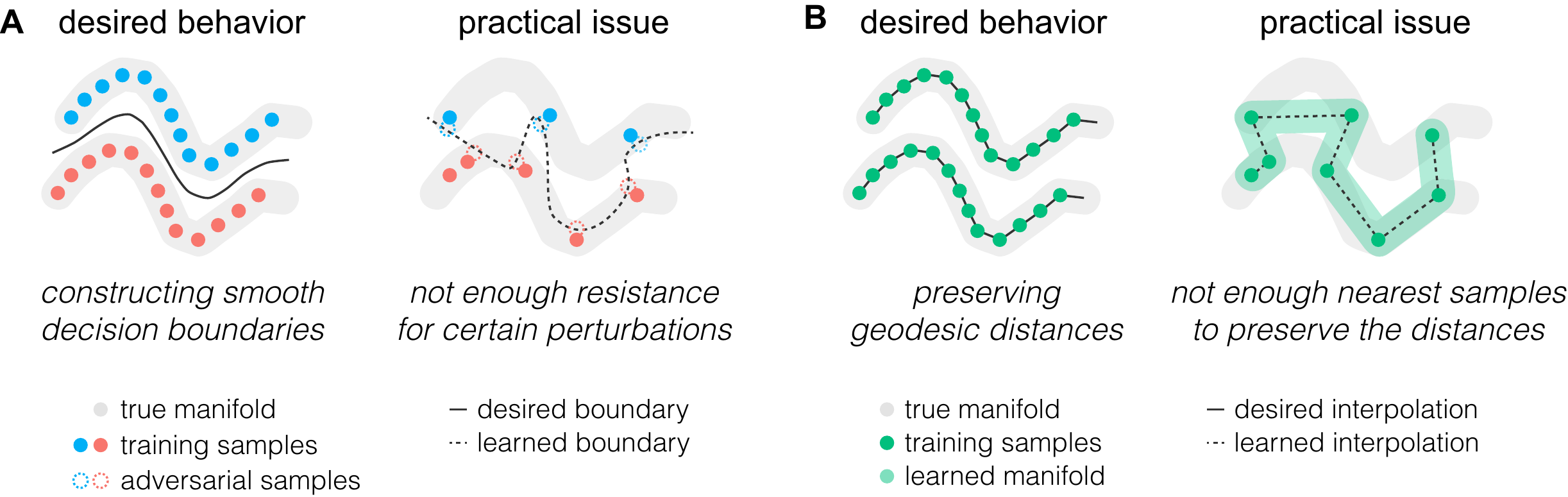}
\caption{Desired behaviors and practical issues of deep learning and manifold learning. (a) Deep learning discriminates different classes; however, it may result in wiggly boundaries vulnerable to adversarial perturbations. (b) Manifold learning preserves geodesic distances; however, it may result in poor embedding.}
\label{f-motivation}
\end{figure}

Previous regularization techniques involve designing efficient training schemes [\ie, dropout~\citep{Srivastava14}, DropConnect~\citep{Wan13}, and Batch Normalization~\citep{Ioffe15}] and well-structured networks [\ie, Network In Network~\citep{Lin13}, and Inception~\citep{Szegedy15}]. Even with these cutting-edge techniques, deep neural networks are still prone to performance degradation when certain small perturbations are injected to samples. The perturbations, which are barely perceptible to humans but make neural networks easily less confident, are called \textit{adversarial examples}~\citep{Szegedy14,Goodfellow15,Nguyen15}. This phenomenon occurs due to fewer training examples than parameters and the inner products of high-dimensional vectors~\citep{Goodfellow15}. In the case of fully connected layers, activation grows by $\epsilon n$ in the worst-case, when $n$ components of an inner product are changed by $\epsilon$.  As shown in Fig.~\ref{f-motivation}(a), decision boundaries constructed by deep networks are wiggly and sensitive to adversarial perturbations.

In~\cite{Goodfellow15} in particular, adversarial training was proposed to minimize classification loss both on given samples and adversarial examples. This type of training helps neural networks to increase the confident scores of corrupted samples and generalize across different clean examples. However, the gradients of a discriminative objective function may vanish when the gradients of both the original and adversarial examples are aggregated.

To resolve this issue, we present a new network called \textit{manifold regularized networks (MRnet)}, which regularizes deep neural networks based on the concept of manifold learning. Manifold learning refers to methodologies based on the manifold hypothesis in which nearest samples in a high-dimensional input space are also nearest pairs on a manifold of much lower dimensionality.

As traditional manifold learning [\ie, ISOMAP~\citep{Tenenbaum00}, Locally Linear Embedding~\citep{Roweis00}, and t-SNE~\citep{Van08}] are well-formulated to find transformation preserving geodesic distances in high-dimensional space, they demonstrate limited performance in practice because of insufficient nearest training samples (Fig.~\ref{f-motivation}(b)). There have been many attempts~\citep{Reed14,Tomar14,Yuan15} to unify deep learning and manifold learning. These studies have common limitations inherited from those of the original manifold learning.

In this paper, we incorporate an explicit penalty term to conserve neighborhood relationships into deep neural networks. With the penalty term, we can preserve neighborhood relationships, which is a objective of manifold learning. Here, we address common limitations of deep networks and manifold learning. We generate adversarial examples which are close enough training samples, and then make neural networks insensitive to adversarial perturbations by embedding the adversarial's near to original samples. Similar to popular deep learning techniques, the manifold penalty can be applied easily to gradient-based optimization. We demonstrate that our method can not only improve classification performances but also find appropriate manifold representations for three benchmark datasets. Especially, the low-dimensional embedding results show effective representations than alternatives.

\section{Related Work}

\subsection{Deep Manifold Learning}

Despite the recent success of deep networks, their ultimate goal has not yet been reached, which is developing general features to solve complex learning tasks. Researchers anticipate that neural networks with multiple layers could learn a manifold embedding and classifier simultaneously~\citep{Bengio13}. However, traditional loss terms, such as a reconstruction or classification error, are not sufficient to capture local variations on manifolds. To support the manifold hypothesis, we need to employ another type of cost function that makes neighborhoods have a similar representation. Hence, there have been many attempts to combine both functional concepts of deep and manifold learning.

Attempts to unify deep and manifold learning can be divided into two categories: manifold learning-based unifying and deep learning-based unifying. The former finds a hierarchical manifold embedding with layer-wise manifold learning. For example, Locally Linear Embedding~\citep{Roweis00} was used as a base unit for a deep architecture~\citep{Yuan15}. The latter incorporates an additional objective from the manifold learning perspective into a framework of deep learning. For instance, the following objective was proposed in~\cite{Reed14}:
\begin{equation}
\sum_{(h_i, h_j) \in \mathcal{D}_{sim}} \Vert h_i - h_j \Vert_2^2 + \sum_{(h_i, h_k) \in \mathcal{D}_{dis}} \max ( 0, \beta - \Vert h_i - h_k \Vert_2 )^2,
\end{equation}
where $\{h_1, \ldots, h_n\}$ is a set of hidden representations obtained from forward operations of deep neural networks. Two sets $\mathcal{D}_{sim}$ and $\mathcal{D}_{dis}$ denote the sets of data pairs with the same labels and different labels, respectively. Another form is defined in~\cite{Tomar14} as follows: $\sum_{\mathcal{D}_{sim}} w_{ij} \Vert h_i - h_j \Vert^2  - \sum_{\mathcal{D}_{dis}} w_{ik} \Vert h_i - h_k \Vert^2$ where $w_{ij} = \exp ( -\Vert x_i - x_j \Vert^2 / \rho )$. Other variations [\ie, \cite{Lu15}, \cite{Bengio04}, and \cite{Rifai11}] are also possible; however, variations still have limitations inherited from the original manifold learning. Training set neighborhood information may be problematic because most nearest samples have too little in common in high-dimensional Euclidean spaces~\citep{Bengio13}.


\subsection{Adversarial Examples}

\begin{figure}
\centering
\psfrag{c}[][][.8]{$x$}
\psfrag{e}[][][1.1]{$\substack{x+ \\ \epsilon \mathrm{sign}( \nabla_x J(\btheta, x, y) )}$}
\includegraphics[width=.75\linewidth]{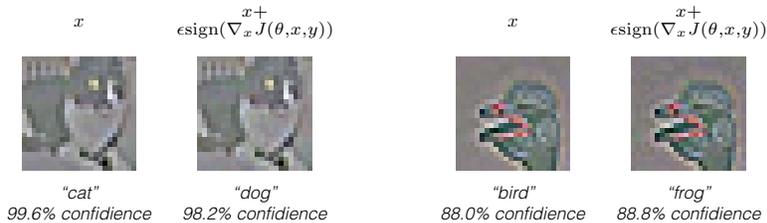}
\caption{Examples of adversarial perturbations on preprocessed CIFAR-10~\citep{Krizhevsky09}.}
\label{f-adv}
\end{figure}

In our study, we incorporate an explicit loss term to preserve neighborhood relationships into deep neural networks. As mentioned earlier, neighborhood information based on only training samples may cause inappropriate embedding. Because the samples are sufficiently densely populated in most cases, we generate adversarial examples that are sufficiently closed to original samples on manifolds. An example of an adversarial perturbation is shown in Fig.~\ref{f-adv} [retrieved from~\cite{Goodfellow15}].

We can obtain these adversarial perturbations easily by updating a sample instead of parameters. While the parameter updating performs $\btheta := \btheta - \eta\nabla_{\btheta} J(\btheta; x, y)$, the sample updating calculates $x := x + \eta\nabla_x J(\btheta; x, y)$, where $\eta$ is a learning rate. The opposite signs in the two update equations mean that the former minimizes a classification loss but the latter maximizes a classification loss. Given a panda image with a high confidence score (see Fig.~\ref{f-adv}), the direction can be calculated by making the neural network less confident. We can produce a new image that a human cannot distinguish from the original, but the neural network believes the image is a gibbon with 99.3\% confidence.

The problem of adversarial perturbations arises in several learning models as well as state-of-the-art deep networks, such as AlexNet~\citep{Krizhevsky12} or GoogLeNet~\citep{Szegedy15}. An ensemble of different deep architectures trained on different subsets of the training data also misclassify the same adversarial example. This suggests that adversarial examples expose fundamental blind spots in our objective functions~\citep{Szegedy14}. Thus, we need to make neural networks resist directions of adversarial perturbations by introducing an explicit loss term to minimize differences between original and adversarial samples.

\section{Manifold Regularized Networks}

Given a data set $\bX=\{ x_1, \ldots, x_n \}$ with corresponding labels $\by=\{ y_1, \ldots, y_n \}$, we consider the following objective loss function:
\begin{equation}
J (\btheta; \bX, \by) = L (\btheta; \bX, \by) + \lambda \Omega(\btheta)
\end{equation}
where $L(\cdot)$ is a classification loss and $\Omega(\cdot)$ is defined to maximize a prior $p(\btheta)$. For the two terms $L(\cdot)$ and $\Omega(\cdot)$, the cross entropy and L2 weight decay are most popular choices in a supervised setting. L2 decay works well in practice; however, these forms of neural networks have intrinsic blind spots due to the huge number of parameters and linear functions using them. Hence, we propose an additional manifold loss term that exploits the characteristics of blind spots.

Suppose we have a neural network with $L+1$ layers. The last $(L+1)$-th layer is an softmax layer. Let $l \in \{ 1, \ldots, L+1\}$ be a layer index and $a^{(l)}$ an activation of the $l$-th layer ($a^{(1)} = x$). The proposed objective can be defined as:
\begin{align}
J_m (\btheta; \bX, \by) = &\ L (\btheta; \bX, \by) + \lambda \Omega(\btheta) + \lambda_m \Phi(\bX, \bX'), \label{e-obj}\\
\Phi(\bX, \bX') = &\ \frac{1}{n} \sum_n \Phi(x_n, x'_n) = \ \frac{1}{2n} \sum_n \Vert a^{(L)}_n - a'^{(L)}_n \Vert_2^2, \\
x'_n = &\ x_n + \beta \underbrace{ \nabla_{x_n} L(\btheta; x_n, y_n) / \Vert \nabla_{x_n} L(\btheta; x_n, y_n) \Vert }_{\Delta x_n} \label{e-adv}
\end{align}
where $\Phi$ is a manifold loss and $\lambda_m$ is a hyper-parameter for the manifold loss. Both $a^{(L)}_n$ and $a'^{(L)}_n$ are the last hidden layer's activations using a standard feed-forward operation. Equation~\eqref{e-adv} denotes the generation of an adversarial example $x'_n$ from $x_n$.

\begin{figure}
\centering
\psfrag{v}[Bl][][.8]{$x, x'$}
\psfrag{w}[Bl][][.8]{$\Phi (x, x')$}
\psfrag{z}[Bl][][.8]{$L (\btheta; x, y)$}

\psfrag{a}[Bl][][.8]{$a^{(1)}$}
\psfrag{c}[Bl][][.8]{$a^{(2)}$}
\psfrag{e}[Bl][][.8]{$a^{(L)}$}
\psfrag{m}[Bl][][.8]{$a^{(L+1)}$}

\psfrag{b}[Bl][][.8]{$\nabla_{\btheta^{(1)}} L (\btheta; x, y)$}
\psfrag{f}[Bl][][.8]{$\nabla_{\btheta^{(L-1)}} L (\btheta; x, y)$}
\psfrag{i}[Bl][][.8]{$\nabla_{\btheta^{(L)}} L (\btheta; x, y)$}
\psfrag{g}[Bl][][.8]{$\delta^{(2)}$}
\psfrag{h}[Bl][][.8]{$\delta^{(L)}$}
\psfrag{k}[Bl][][.8]{$\delta^{(L+1)}$}
\psfrag{l}[c][][.8]{$\nabla_x L(\btheta; x, y)$}

\psfrag{n}[Bl][][.8]{$a'^{(1)}$}
\psfrag{o}[Bl][][.8]{$a'^{(2)}$}
\psfrag{r}[Bl][][.8]{$a'^{(L)}$}
\psfrag{s}[Bl][][.8]{$a'^{(L+1)}$}

\psfrag{j}[Bl][][.8]{$\delta_1^{(2)},\ \delta_2^{(2)}$}
\psfrag{p}[Bl][][.8]{$\delta_1^{(L)},\ \delta_2^{(L)}$}

\psfrag{x}[Bl][][.8]{$\nabla_{\btheta^{(L-1)}} \Phi (x, x')$}
\psfrag{u}[Bl][][.8]{$\nabla_{\btheta^{(1)}} \Phi (x, x')$}
\includegraphics[width=\linewidth]{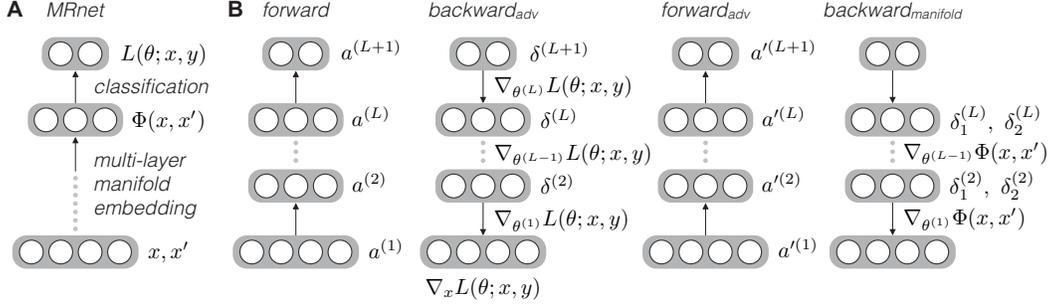}
\caption{(a) The proposed methodology learns both classifier and manifold embedding that is robust for adversarial perturbations. (b) Forward and backward operations of MRnet. The first forward operation is the same as in a standard neural network. The following $\mbox{backward}_{adv}$ is the same as the standard back-propagation except that an adversarial perturbation $\nabla_x L(\btheta; x, y)$ is computed in the first layer.}
\label{f-proposed}
\end{figure}

Fig.~\ref{f-proposed} shows an overview of the proposed methodology. To make a deep neural network robust to adversarial perturbations, we regard the activations of the last hidden layer as a manifold representation and minimize the difference between the two manifold embeddings of $x$ and $x'$ as shown in Fig.~\ref{f-proposed}(a). The proposed forward and backward operations to solve~\eqref{e-obj} are presented in Fig.~\ref{f-proposed}(b). The forward operation is the same as in a standard neural network:
\begin{align}
a^{(l+1)} = &\ f(z^{(l+1)}) = f(W^{(l)} a^{(l)} + b^{(l)}), &\mbox{if } l \mbox{-th layer is fully-connected}, \\
\big[ a^{(l+1)} \big]_j = &\ f( \big[ z^{(l+1)} \big]_j ) = f(\sum_i \big[ a^{(l)} \big]_i * W_{ij}^{(l)}+ b_j^{(l)}), &\mbox{if } l \mbox{-th layer is convolutional}
\end{align}
where $f(\cdot)$ is an activation function (chosen to be the rectified linear unit in this paper) and $*$ is the convolutional operation of a valid size. For a convolutional layer, $[a^{(l)}]_i$ and $[a^{(l+1)}]_j$ are the $i$-th and $j$-th feature map in the $l$-th layer and ($l$+$1$)-th layer, respectively. $W_{ij}^{(l)}$ is a convolutional filter between $[a^{(l)}]_i$ and $[a^{(l+1)}]_j$. The following backward$_{adv}$ is also the same as standard back-propagation except that an adversarial perturbation $\nabla_x L(\btheta; x, y)$ is computed in the first layer:
\begin{align}
\nabla_{x} L(\btheta; x, y) = &\ \big( \delta^{(2)} \circ g(a^{(2)}) \big) W^{(1)}, &\mbox{if } 1 \mbox{-st layer is fully-connected}, \\
\Big[\nabla_{x} L(\btheta; x, y)\Big]_i = &\ \sum_j \big( \big[ \delta^{(2)} \big]_j \circ g( \big[ a^{(2)} \big]_j ) \big) \star W_{ij}^{(1)}, &\mbox{if } 1 \mbox{-st layer is convolutional}
\end{align}
where $g(\cdot)$ is the derivative of an activation function, $\star$ is the convolutional operation of the full size, and $\circ$ is the Hadamard product (an element-wise multiplication).

Next, another forward operation is performed to obtain all the activations of an adversarial example $a'^{(1)}=x'=x+\Delta x$ and calculate $\nabla_{\btheta} \Phi (\bX, \bX')$. Recall that $\Phi (x_n, x_n')$ is the squared error between the original $a^{(L)}_n$ and its adversarial $a'^{(L)}_n$:
\begin{equation}
\Phi (x_n, x'_n) = (1/2) \Vert a^{(L)}_n - a'^{(L)}_n \Vert_2^2 = (1/2) ( a^{(L)}_n - a'^{(L)}_n )^T ( a^{(L)}_n - a'^{(L)}_n ).
\end{equation}
Similar to the backpropagation algorithm, we first compute error terms $\delta^{(L)}_1$ and $\delta^{(L)}_2$:
\begin{equation}
\delta^{(L)}_1 = -( a'^{(L)} - a^{(L)} ), \quad \delta^{(L)}_2 = -( a^{(L)} - a'^{(L)} ).
\end{equation}
The error $\delta^{(L)}_1$ is the difference of $a'^{(L)}$ with respect to $a^{(L)}$, and the error $\delta^{(L)}_2$ is the difference of $a^{(L)}$ with respect to $a'^{(L)}$. Now, we can present the gradients and back-propagation for the manifold loss term $\Phi$ for each layer $l=L, \ldots, 2$. If the $l$-th layer is fully-connected, the rules are as follows:
\begin{align}
\nabla_{W^{(l-1)}} \Phi = &\ \frac{1}{n} \Big( \big( \delta^{(l)}_1 \circ g(a^{(l)}) \big)  (a^{(l-1)})^T + \big( \delta^{(l)}_2 \circ g(a'^{(l)}) \big)  (a'^{(l-1)})^T \Big), \\
\nabla_{b^{(l-1)}} \Phi = &\ \frac{1}{n} \Big( \big( \delta^{(l)}_1 \circ g(a^{(l)}) \big)  \bone + \big( \delta^{(l)}_2 \circ g(a'^{(l)}) \big)  \bone \Big), \\
\delta^{(l-1)}_1 = &\ (W^{(l)})^T \big( \delta^{(l)}_1 \circ g(a^{(l)}) \big), \quad \delta^{(l-1)}_2 = \ (W^{(l)})^T \big( \delta^{(l)}_2 \circ g(a'^{(l)}) \big).
\end{align}

The rules for a convolutional layer are as follows:
\begin{align}
\Big[ \nabla_{W^{(l-1)}} \Phi \Big]_i = &\ \frac{1}{n} \sum_j \Big( a^{(l-1)} * \big( \delta^{(l)}_1 \circ g(a^{(l)}) \big) + a'^{(l-1)} * \big( \delta^{(l)}_2 \circ g(a'^{(l)}) \big) \Big), \\
\Big[ \nabla_{b^{(l-1)}} \Phi \Big]_i = &\ \frac{1}{n} \sum_{row} \sum_{col} \sum_{n} \Big( \big( \delta^{(l)}_1 \circ g(a^{(l)}) \big) + \big( \delta^{(l)}_2 \circ g(a'^{(l)}) \big) \Big), \\
\big[ \delta^{(l-1)}_1 \big]_i = &\ \sum_j \big[ \delta^{(l)}_1 \circ g(a^{(l)}) \big]_j \star W^{(l)}_{ij}, \quad \big[ \delta^{(l-1)}_2 \big]_i = \sum_j \big[ \delta^{(l)}_2 \circ g(a'^{(l)}) \big]_j \star W^{(l)}_{ij}.
\end{align}

\section{Experiments}

\begin{figure}
\centering
\includegraphics[width=\linewidth]{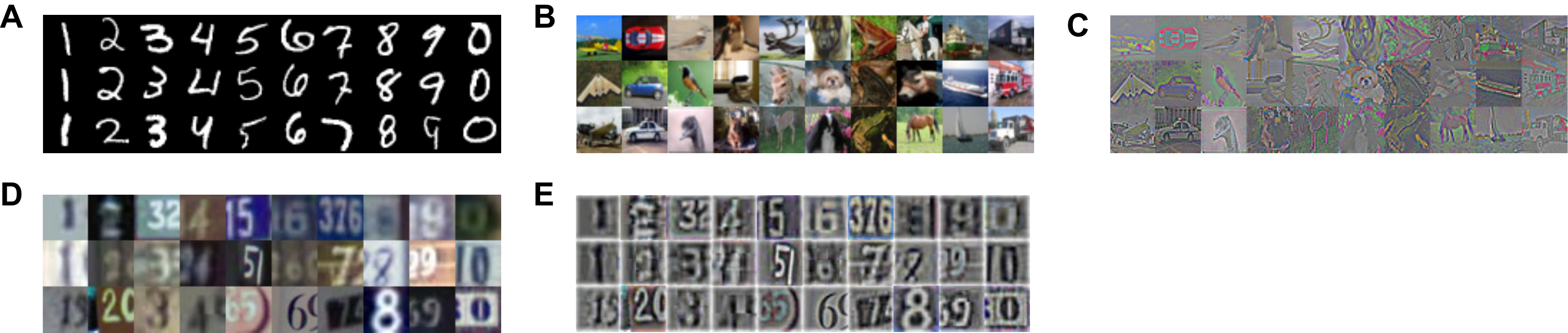}
\caption{Three datasets we tested. (a) The MNIST. (b, c) The rawdata and its normalized version of the CIFAR-10. (d, e) The rawdata and its normalized version of the SVHN. The normalization is performed with local contrast normalization to manipulate the extreme brightness and color variations.}
\label{f-dataset}
\end{figure}

The optimization of the proposed method was conducted using standard stochastic gradient descent. The mini-batch size and the momentum were set to 100 and 0.9, respectively. The learning rate was annealed as described in~\cite{Wan13}. For each subsection, we present an initial learning rate and three numbers of epochs, such as 0.001 (100-20-10). We trained models with the initial rate for the first number of epochs. Then, we multiplied by 0.1 for the second epochs followed by 0.5 again for the third epochs. For example, the learning schedule 0.001 (100-20-10) denotes a learning rate of 0.001 for 100 epochs followed by 0.0001 for 20 epochs and 0.00005 for 10 epochs.

We evaluated the proposed regularization on three benchmark datasets: MNIST~\citep{LeCun98}, CIFAR-10~\citep{Krizhevsky09}, and SVHN~\citep{Netzer11}, as depicted in Fig.~\ref{f-dataset}. In the case of the CIFAR and SVHN, we preprocessed the data using zero-phase component analysis (ZCA) whitening and local contrast normalization; the same techniques as~\cite{Goodfellow13} and~\cite{Zeiler13}, respectively. These preprocessing techniques are known for normalizing the extreme brightness and color variations efficiently.

Our implementation was based on Caffe~\citep{Jia14}, which is one of the most popular deep learning frameworks. We added new forward-backward steps and customized four types of layers: convolutional, pooling, response normalization~\citep{Hinton12}, and fully-connected layers. The codes and all the details of hyper-parameters [\ie, weight decay and the number of hidden nodes] are available on our GitHub page~\footnote{https://github.com/taehoonlee/caffe}. For each case, we presented a mean value with a standard deviation among 10 runs as a format of $\mu \pm \sigma$. For the generation of adversarial examples, we had to maintain an appropriate noise level $\beta$. This is shown in detail in the appendix and Fig.~\ref{f-noiselevel}.

\subsection{Improved classification performance}

\paragraph{MNIST}

The MNIST~\citep{LeCun98} dataset is one of the most popular benchmarks and consists of a set of $28 \times 28$  grayscale images with corresponding labels 0--9. Because these images have high contrast like binary images, typically they have been tested without any pre-processing. To obtain desirable representations in the manifold space, we tested two types of models.

First, we trained models with two fully connected layers each with 800 hidden units, followed by a softmax layer. We used a learning schedule of 0.1 (40-40-20) and obtained a test accuracy of $98.848 \pm 0.052$\%, which is the best record over the published results with only two hidden layers. DropConnect~\citep{Wan13} and dropout~\citep{Srivastava14} produced $98.800 \pm 0.034$\% and $98.720 \pm 0.040$\%, respectively.

Additionally, we conducted MRnet with two convolutional layers followed by two fully connected layers. The results are summarized in Table~\ref{t-mnist-conv}. The top four methods were performed 10 times using our implementation while the results of the bottom five methods were reported in the literature. The previous best published result is the $99.55$\% test accuracy of Maxout~\citep{Goodfellow13}. Among the alternatives compared, MRnet achieved the state-of-the-art discriminative performance: $99.536 \pm 0.045$\% (best: $99.58$\%).


\begin{table}
\caption{Test set accuracy on the MNIST with convolutional hidden layers.}
\label{t-mnist-conv}
\small
\centering
\renewcommand{\arraystretch}{1.1}
\begin{tabular}{lc}
\toprule
Method & Test accuracy (\%) \\
\midrule
\bf{MRnet + dropout} & $99.536 \pm 0.045$ \\
Batch Normalization~\citep{Ioffe15} & $99.465 \pm 0.051$ \\
Dropout~\citep{Srivastava14} & $99.482 \pm 0.053$ \\
Only L2-decay & $99.364 \pm 0.055$ \\
\midrule
DropConnect~\citep{Wan13} & $99.370 \pm 0.035$ \\
Maxout + dropout~\citep{Goodfellow13} & $99.55$ \\
NIN + dropout~\citep{Lin13} & $99.53$ \\
Stochastic Pooling~\citep{Zeiler13} & $99.53$ \\
Lasso in F-layers~\citep{Jarrett09} & $99.47$ \\
\bottomrule
\end{tabular}
\end{table}

\begin{figure}
\centering
\psfrag{a}[][][.8]{$\Vert a^{(L)} - a'^{(L)} \Vert_2$}
\includegraphics[width=.32\linewidth]{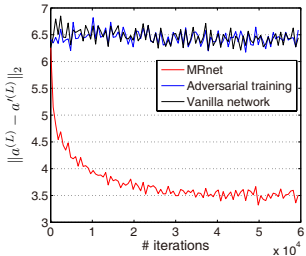}
\caption{Difference between $a^{(L)}$ and $a'^{(L)}$ over iterations.}
\label{f-manifolddistance}
\end{figure}

\paragraph{CIFAR-10}

The CIFAR-10~\citep{Krizhevsky09} dataset is a set of $32 \times 32$ color images of 10 classes. There are 50,000 training and 10,000 test images. We applied ZCA preprocessing ($\epsilon=0.01$) similar to~\cite{Goodfellow13}. To be consistent with previous work, we evaluated our method with $24 \times 24$ random cropping and horizontal flipping augmentation. The learning schedule was set to 0.001 (100-10-10).

The results of Caffe runs and the literature are summarized in Table~\ref{t-cifar10}. Note that a locally connected layer in the description column is a weight-unshared convolutional layer. We can achieve a test accuracy of $91.082 \pm 0.237\%$, which sets a new performance record.

\begin{table}
\caption{Test set accuracy on the CIFAR-10 with data augmentation and the SVHN.}
\label{t-cifar10}
\small
\centering
\renewcommand{\arraystretch}{1.1}
\begin{tabular}{llcc}
\toprule
\multirow{2}{*}{Method} & \multirow{2}{*}{Description} & \multicolumn{2}{c}{Test accuracy (\%)} \\
 &  & CIFAR-10 & SVHN \\
\midrule
\bf{MRnet + dropout} & 4C$^{1}$ + 2F$^{2}$ & $ 91.182 \pm 0.237 $ & $97.521 \pm 0.052$ \\
MRnet + dropout & 3C + 2F & $ 89.835 \pm 0.224 $ & $97.327 \pm 0.048$ \\
Adv training + dropout~\citep{Goodfellow15} & 3C + 2F & $ 89.837 \pm 0.219 $ & $97.323 \pm 0.046$ \\
Batch Normalization~\citep{Ioffe15} & 3C + 2F & $ 89.032 \pm 0.214 $ & $97.111 \pm 0.047$ \\
Dropout~\citep{Srivastava14} & 3C + 2F & $ 89.097 \pm 0.240 $  & $ 97.058 \pm 0.086 $ \\
Only L2-decay & 3C + 2F & $ 88.259 \pm 0.225$ & $ 96.897 \pm 0.058$ \\
\midrule
DSN~\citep{Lee15} & - & $ 91.78 $ & $ 98.08 $ \\
DropConnect~\citep{Wan13} & 2C + 2L$^{3}$ + 1F & $ 90.59 $ & $ 97.77 \pm 0.039 $ \\
Maxout + dropout~\citep{Goodfellow13} & 3M$^{4}$ + 1F & $90.62$ & $97.53$ \\
NIN + dropout~\citep{Lin13} & 3N$^{5}$ + 1F & $91.19$ & $97.65$ \\
Stochastic Pooling~\citep{Zeiler13} & 3C & (no aug) $84.87$ & $97.20$ \\
\bottomrule
\multicolumn{4}{l}{$^{1}$a convolutional layer, \quad $^{2}$a fully connected layer, \quad $^{3}$a locally connected layer} \\
\multicolumn{4}{l}{$^{4}$a maxout convolutional layer, \quad $^{5}$a network in network convolutional layer}
\end{tabular}
\end{table}

\paragraph{SVHN}

The SVHN~\citep{Netzer11} dataset is composed of 604,388 training and 26,032 test images of 10 classes. Following~\cite{Zeiler13}, we conducted local contrast normalization with a $3 \times 3$ filter and $28 \times 28$ random cropping. The structure and parameters used in the SVHN are the same as those used for the CIFAR-10, which consists of three or four convolutional layers followed by two fully connected layers. For this dataset, we obtained a test accuracy of $97.521 \pm 0.052$ with a learning schedule of 0.001 (20-20-20). A summary with the alternatives is provided in Table~\ref{t-cifar10}.


\subsection{Disentanglement and generalization}

The proposed manifold loss $\Phi$ minimizes the 2-norm difference between $a^{(L)}$ and $a'^{(L)}$. In other words, the difference in the last hidden layer's activations between an original and its adversarial sample is minimized. However, an adversarial perturbation at the beginning of training is meaningless. Thus, we applied $\Phi$ after pre-training a vanilla neural network with several iterations. Fig.~\ref{f-manifolddistance} shows the variation of $\Vert a^{(L)} - a'^{(L)} \Vert_2$ over the number of iterations. The iteration 0 denotes the point at which $\Phi$ is applied.

As illustrated in Fig.~\ref{f-manifolddistance}, MRnet minimized the manifold distances more successfully than a vanilla network. The final value of $\Vert a^{(L)} - a'^{(L)} \Vert_2$ was 3.57. In the network we tested, the number of hidden nodes in the last hidden layer was 1024, and the average difference of individual activation values can be calculated approximately as $0.1\ (=\sqrt{3.57^2 / 1024} )$. Because each activation value in the last hidden layer ranged from 0 to 2.74, the average difference 0.1 means that $a^{(L)}$'s and $a'^{(L)}$'s were projected into very closed regions by the proposed multi-layer manifold embedding.

\begin{figure}
\centering
\includegraphics[width=\linewidth]{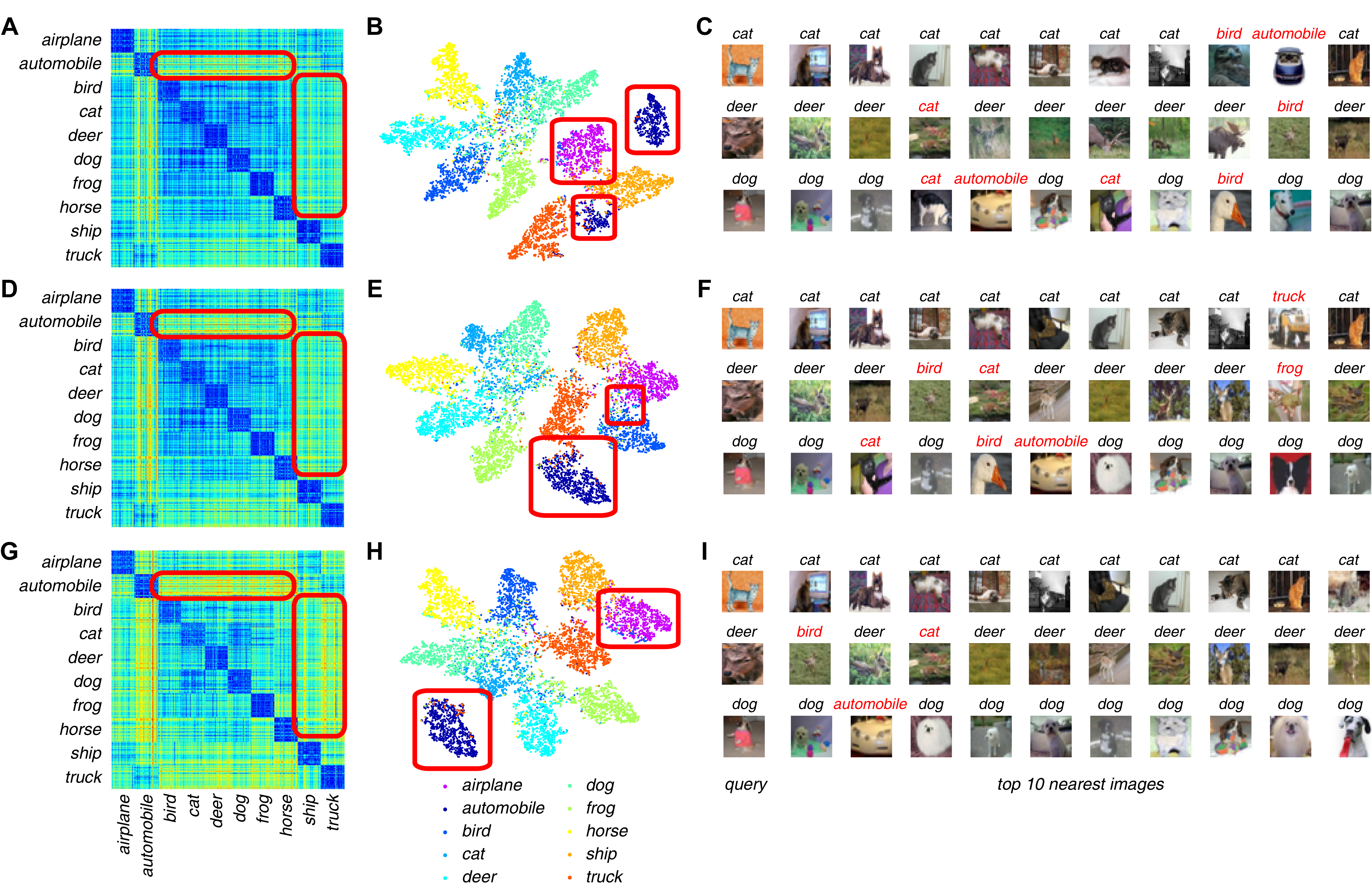}
\caption{Embedding results of the CIFAR-10 test set. (a) Pairwise distance matrix of $a^{(L)}$ from a \textit{vanilla network}. (b) 2-D visualization of the manifold embedding through t-SNE from a \textit{vanilla net}. (c) Query images and top 10 nearest images from a \textit{vanilla net}. Distance matrices, t-SNE plots, and query with the nearest images from (d-f) \textit{adversarial training} and (g-i) \textit{MRnet}.}
\label{f-manifold}
\end{figure}

We also visualized $a^{(L)}$ of the CIFAR-10 test set to examine the effect of the proposed manifold loss $\Phi$. Fig.~\ref{f-manifold}(a-c) and Fig.~\ref{f-manifold}(d-f) show the embedding results without and with $\Phi$, respectively. As shown in Fig.~\ref{f-manifold}(d), $\Phi$ increases the contrast of pairwise distances between intra-classes (block-diagonal elements) and inter-classes (other elements) compared with Fig.~\ref{f-manifold}(a). The contrast can be quantified using two clustering evaluation metrics: the silhouette coefficients $\in [-1, 1]$~\citep{Rousseeuw87} and the Dunn index~\citep{Dunn73}. As clusters are separated better, both metrics produce higher values. Without and with $\Phi$, the average values of silhouette coefficients do not exhibit significant differences. However, the Dunn indices without and with $\Phi$ were 0.0297 and 0.0433, respectively. Because the denominator of the Dunn index is the maximum distance within a cluster~\citep{Dunn73}, we argue the Dunn index can present more appropriate quantification of the contrast, and obtained 31.45\% improved contrast.

A similar effect can be found in the 2-D visualization with t-SNE~\citep{Van08} [see Fig.~\ref{f-manifold}(b) and (e)]. Without $\Phi$, two boxed chunks of the automobile class are separated and one is closed to the ship and truck classes. However, $\Phi$ can make instances of the automobile class be grouped into one chunk as depicted in Fig.~\ref{f-manifold}(e). Finally, Fig.~\ref{f-manifold}(c) and Fig.~\ref{f-manifold}(f) present a few query images and the top 10 nearest neighbors on the manifold embedding space. The 10-th closest bird image of the dog and the 9-th closest bird image of the deer were eliminated in the list of nearest neighbors after applying $\Phi$.


\section{Conclusion}

We have proposed a novel methodology, unifying deep learning and manifold learning, called manifold regularized networks (MRnet). Traditional neural networks, even state-of-the art deep networks, have intrinsic blind spots due to a huge number of parameters and linear function components using them. We tested MRnet and confirmed its improved generalization performance underpinned by the proposed manifold loss term on deep architectures. By exploiting the characteristics of blind spots, the proposed MRnet can be extended to the discovery of true representations on manifolds in various learning tasks.

\bibliography{mybib}
\bibliographystyle{iclr2016_conference}

\clearpage

\section*{Appendix}

\subsection*{Generation of Adversarial Examples}

For the generation of adversarial examples, we used constant-norm perturbation, while the max-norm constrained perturbation was used in the original paper regarding the problems of adversarial examples. In our experiments, the constant-norm and max-norm approaches did not exhibit significant differences. The important aspect was to select a degree of adversarial perturbations $\beta$.

\begin{figure}
\centering
\psfrag{a}[][][.7]{$\beta = 0$}
\psfrag{c}[][][.7]{$1$}
\psfrag{e}[][][.7]{$2$}
\psfrag{m}[][][.7]{$3$}
\psfrag{b}[][][.7]{$\beta = 0$}
\psfrag{d}[][][.7]{$100$}
\psfrag{f}[][][.7]{$200$}
\psfrag{h}[][][.7]{$300$}
\psfrag{n}[][][.7]{$\beta = 0$}
\psfrag{o}[][][.7]{$100$}
\psfrag{r}[][][.7]{$200$}
\psfrag{s}[][][.7]{$300$}
\includegraphics[width=\linewidth]{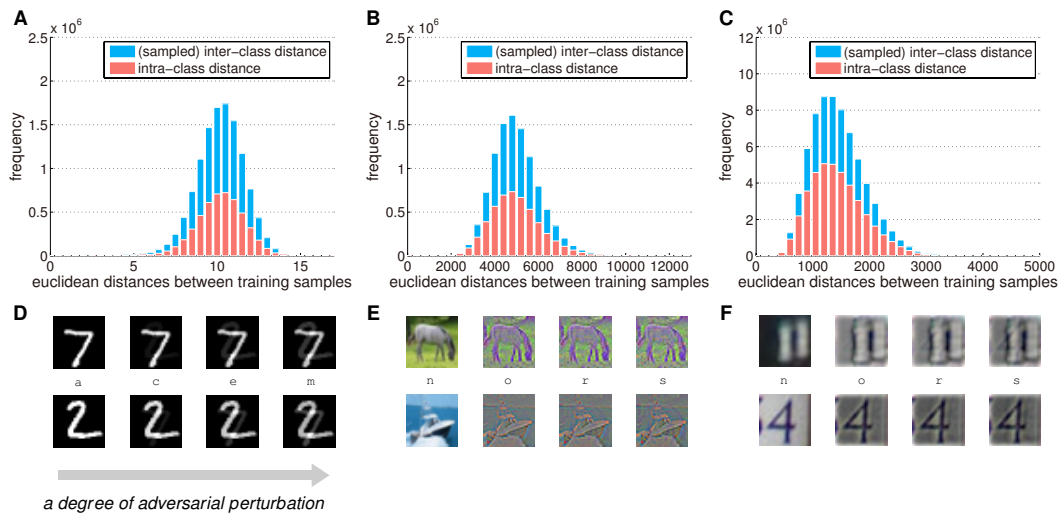}
\caption{(a,b,c) Distributions of Euclidean distances between training samples on individual datasets. (d,e,f) Different perturbation levels on individual datasets. Note that $x'=x + \beta \Delta x$ where $\Delta x$ is an adversarial perturbation of a unit norm ($\norm{\Delta x}_2=1$). We chose $\beta$ in the range that did not violate class information.}
\label{f-noiselevel}
\end{figure}

Because adversarial examples become actual instances of a different class when $\beta$ is greater than a certain threshold, we had to select a degree of perturbation level $\beta$ carefully. In Fig.~\ref{f-noiselevel}(a-c), each histogram shows the distribution of pairwise distances. We set $\beta$ in three datasets to 2, 200, and 200, respectively. These noise levels are small enough to maintain class memberships and the manifold hypothesis. We recommend $\beta$ in the range of minimum values among intra-class distances.

\end{document}